\documentclass[a4paper, 10pt, conference]{ieeeconf}      
\IEEEoverridecommandlockouts                              
\usepackage{graphicx}
\usepackage{amsmath}
\usepackage{amssymb}
\usepackage{comment}
\usepackage{hyperref}       
\usepackage[utf8]{inputenc}
\usepackage{datetime}
\usepackage{url}
\hypersetup{
    colorlinks=true,
    linkcolor=blue,
    filecolor=magenta,      
    urlcolor=cyan,
}

\usepackage{subcaption}

\usepackage{mathtools}
\DeclarePairedDelimiter\floor{\lfloor}{\rfloor}

\title{\LARGE \bf
Design and implementation of an environment \\ for Learning to Run a Power Network (L2RPN)
}

\usepackage{todonotes}

\newcommand{\isabelle}[1]{\todo[inline,color=orange!40]{#1 -- Isabelle}}

\newcommand{\benjamin}[1]{\todo[inline,color=green!40]{#1 -- Benjamin}}
\newcommand{\marvin}[1]{\todo[inline,color=pink!40]{#1 -- Marvin}}

\author{Marvin Lerousseau$^{1, 2}$ \\ $^{1}$ Grenoble INP - Ensimag $^{2}$ Université Grenoble Alpes\\ June 17, 2018}

\begin{document}

\maketitle

\pagestyle{plain}

\begin{abstract}

This report summarizes work performed as part of an internship at INRIA, in partial requirement for the completion of a master degree in math and informatics. The goal of the internship was to develop a software environment to simulate electricity transmission
in a power grid and actions performed by operators to maintain this grid in security. Our environment lends itself to automate the control of the power grid with reinforcement learning (RL\cite{sutton}) agents, assisting
human operators. It is amenable to organizing benchmarks, including a challenge in machine learning planned by INRIA and RTE for 2019. Our framework, built on top of open-source libraries, is available at \url{https://github.com/MarvinLer/pypownet}.
In this report we present intermediary results and its usage in the context of a Reinforcement Learning (RL) game.

\end{abstract}

\section{INTRODUCTION}
This project addresses technical aspects related to the transmission of electricity in extra high voltage and high voltage power networks (63kV and above),
such as those managed by the company RTE (R\'eseau de Transport d'Electricit\'e) the French TSO (Transmission System Operator).
Numerous improvements of the efficiency of energy infrastructure are anticipated in the next decade from the deployment of smart grid technology in power distribution networks to more accurate consumptions preditive tools.
As we progress in deploying renewable energy harnessing wind and solar power to transform it to electric power, we also expect to see growth in power demand for new applications such as electric vehicles.
Electricity is known to be difficult to store at an industrial level.
Hence supply and demand on the power grid
must be balanced at all times to the extent possible. Failure to achieve this balance may result in network breakdown and subsequent power outages of various levels of gravity.
Principally, shutting down and restarting power plants (particularly nuclear power plants) is very difficult and costly since there is no easy way to switching on/off generators.
Many consumers, including hospitals and people hospitalized at home as well as factories critically suffer from power outages.
Using machine learning (and in particular reinforcement learning) may allow us to optimize better the operation of the grid eventually leading to reduce redundancy in transmission lines, and make better utilization of generators, and lower power prices. 
The goal of this project is to prepare a data science challenge to engage the scientific community to help solving
this difficult problem.

RTE is a Transmission System Operators (TSO).
One of the objective of TSOs is to route electricity from productions to consumptions using a power grid, under the constraint of avoiding equipment failure. A typical failure we are interested in is line unplanned outage resulting from over-heating.
Such incidents happen when lines are subject to power flows greater than a nominal threshold value. To avoid line failures (and possible subsequent cascading failures), operators (dispatchers) have a set of actions at disposal: they can locally modify the lines interconnections, switch on or switch off power lines, or change electricity production. 

The difficulty of the task arises from the complexity of the network architecture, also called grid topology, which frequently changes due to events such as hardware failures (e.g. due to weather conditions such as thunderstorms), planned maintenance or preventive actions. On top of that, rising renewable energies are less predictable than conventional productions systems (e.g. nuclear plants), bringing more uncertainty to the productions schemes. In this context, we are interested in developing tools that will assist dispatchers to maintain a power grid safe and face the increasing complexity of their task. This work builds on top of work performed by Benjamin Donnot\cite{donnot1}\cite{donnot2}\cite{donnot3} as part of is PhD thesis and Joao Araujo, an intern having performed preliminary work on the subject last summer.

Recent work in deep learning\cite{deeplearningbook} has underlined the potential of deep neural networks in solving complex tasks (\cite{dlex}, \cite{dlex2}).
For classification and regression tasks, they are usually trained using supervised learning, which necessitates a labeled dataset. In our case, a suitable dataset could be made of pairs of grid situations and dispatchers curative actions, such that the models are trained by copying (and hopefully generalizing) the dispatchers actions given a grid state (or a temporal chronic of grids photos).
Unfortunately, we do not have access to such labeled data providing preventive or remedial actions of dispatchers for given crisis situations, for very-high voltage grid\footnote{All of the actions of the dispatchers are recorded and without proper annotations so their motivations are not accurately documented. Besides, a lot of these actions are anticipative, which necessitates extra additional amounts of data including consumptions predictions and planned productions.}.

This prompted us to investigate methods of reinforcement learning. Recent papers (\cite{starcraft}, \cite{deepmindalphagozero}, \cite{rlex2})
managed to successfully apply reinforcement learning to high-dimensional temporal tasks. One specific aspect of our problem is that the power grid can be accurately simulated with a physical simulator implementing the laws of physics (ordinary differential equations) under some quite restrictive assumptions, for example that we are in a quasi stationary regime. These hypotheses are quite common in the power system community.
Therefore our problem lends itself well to reinforcement learning because data can be generated using an {\em Environment simulator} (a power grid physical simulator). The hope is that a trained model would implement a policy (mapping states of the network to preventive or curative actions that maintain the network in security over time),
that might be used to assist human dispatchers in making the right decision.
In our project, we simplified the overall problem by limiting ourselves to toy examples of grids and subsets of actions to create a ``serious game'' simulating semi-realistic conditions of power grid control. This game lends itself to reinforcement learning solutions. The proposed framework will be used for a challenge implemented on the Codalab platform (\url{http://competitions.codalab.org}).
\\

This document is organized as follows. First, we give some context about power grids and large-scale power grid conduct. Second, we review recent works about machine learning applied to Power Systems and Reinforcement Learning. We then present our contribution to the design and implementation of the framework. The following part discusses some early results obtained using the proposed game. Finally, the last part lists key elements of our future work.

\section{BACKGROUND}
\subsection{Power grids}
\begin{figure}[ht!]
\centering
\includegraphics[width=\linewidth]{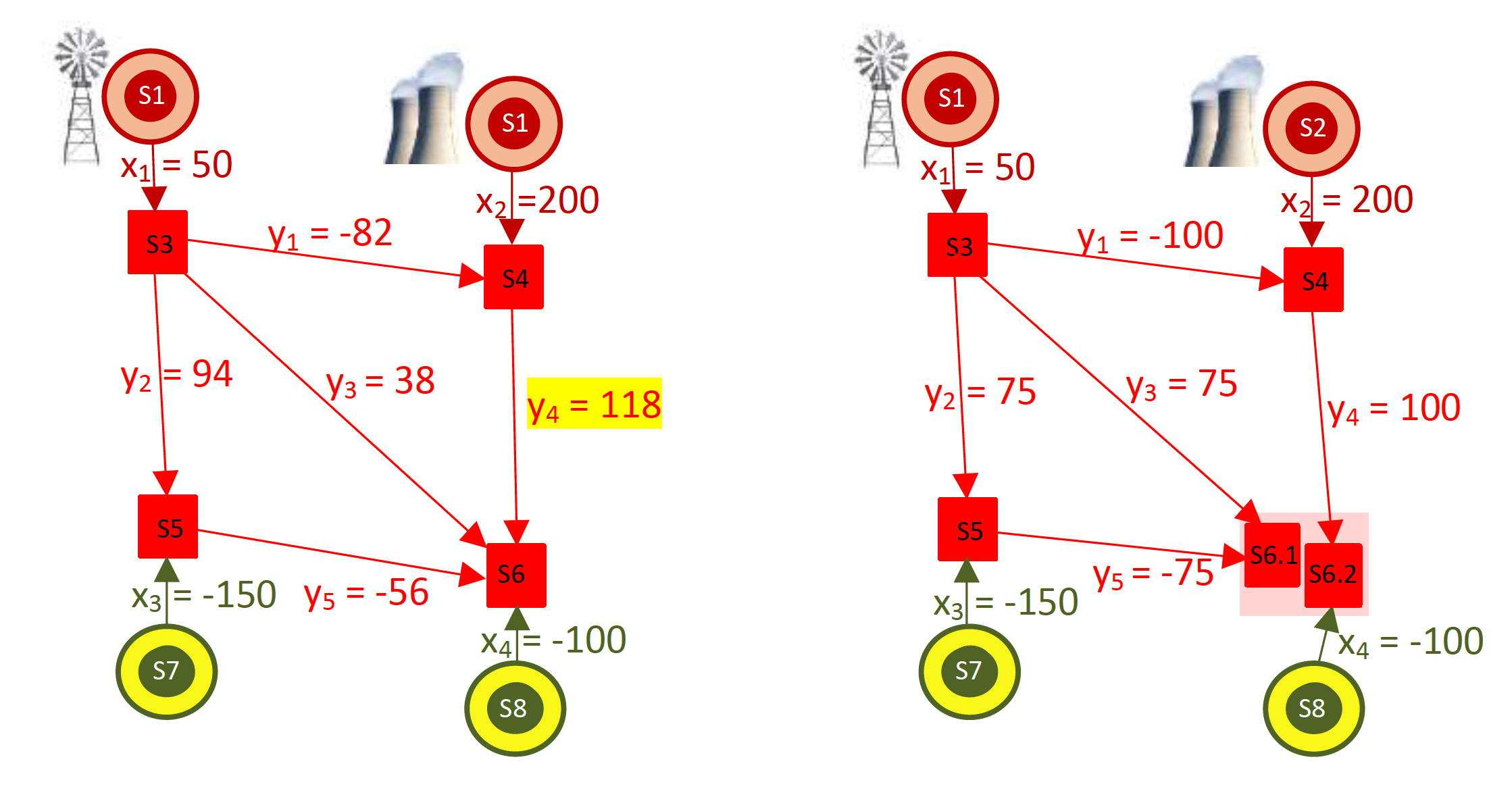} 
(a)~~~~~~~~~~~~~~~~~~~~~~~~~~~~~~~~~~(b)
\caption{{\bf Mini example of power transmission grid.} Electricity must be transported from production nodes (brown circles) to consumption nodes (green circles). They are interconnected through a network (grid) of transmission lines (red lines), connected at substations (red squares). The injections $X=(x_1, x_2, x_3, x_4)$, which include both productions and consumptions, must add up to zero. The way in which lines are interconnected is referred to as grid topology $\tau$. The flows in the red lines $Y$ result from the injections and the topology: $Y = S(X, \tau)$. At any time, the grid operators (or {\em dispatchers}) must make sure that the network is operated in security and no line exceeds its thermal limit (a current flow above which the line might melt). (a) Line $y_4$ goes over its thermal limit 100. (b) A change in topology (splitting of node 6) brings $y_4$ back to its thermal limit.
}
\label{fig:mini_nets}
\vspace{-0.5cm}
\end{figure}

\begin{figure}
\centering
\begin{subfigure}{.25\textwidth}
  \centering
  \includegraphics[width=.5\linewidth]{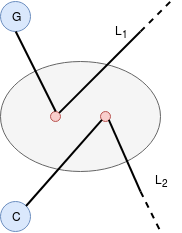}
  \label{fig:config1}
  \caption{Representation a}
\end{subfigure}%
\begin{subfigure}{.25\textwidth}
  \centering
  \includegraphics[width=.5\linewidth]{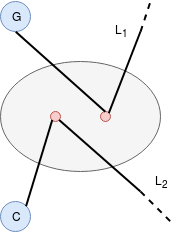}
  \label{fig:config2}
  \caption{Representation b}
\end{subfigure}
\caption{Example of representations of inner configuration of a substation. The substation is the gray ellipse. Buses are depicted the two pink filled circles. Because buses do not have proper physical meaning, both figures represent the same configuration. For this configuration, production G and branch $L_1$ are directly connected, and consumption C and branch $L_2$ are directly connected.}
\label{fig:config}
\end{figure}

A power grid is an network made of electric hardware
and intended to transmit electricity from productions to consumptions. See Fig. \ref{fig:mini_nets} for a representation of a power grid. Formally, the structure of a power grid is a graph $G=\{V, E\}$ with $V$ the set of nodes and $E$ the set of edges. Edges represent the power lines, also called branches or transmission lines. In practice, $V$ is the set of substations, which are physical entities on which other elements can be interconnected, such as productions (e.g. solar farm) or consumptions (such as a city). A bus is a mathematical concept referring to an intersection of directly connected elements within a substation. Substations can have multiple buses, in the sense that elements can be directly wired to any subset of the other elements. Fig. \ref{fig:config} displays two representations of a substation composed of 4 elements: one generator $G$, one consumption $C$, and two power lines $L_1$ and $L_2$. In this particular topological configuration, there are 2 buses: the bus made of $G$ and $L_1$ and the one composed of $C$ and $L_2$. Note that buses do not have proper physical representation: as such, both representations in \ref{fig:config} are strictly equivalent (invariance by bus permutation).

Concretely, substations are made of electric poles on which the connectible elements can be wired. TSOs technicians can change the pole on which an element is connected by operating switches. They can also switch off elements such as branches such that they are temporarily removed from the grid (to repaint power lines for example). Moreover, there cannot be only one element connected to a pole, since the electricity would have no exit point. In this work, we will constraint the substations to have at most two buses, i.e. element of a substation can be grouped into a maximum of two groups, in which objects are directly connected. In other words, the elements of every substation can be interconnected into one or two groups.
\\

\vspace{0.7cm}

On top of the graph structure, a power grid is submitted to Kirchoff's laws. For instance, at any node, the sum of input power
is equal to the sum of output power. Given a set of injections (productions and consumptions values), a network structure and physical laws, a grid will naturally converge to an equilibrium,
also called a steady-state. We are especially interested in the flows circulating in the branches of the grid in steady-states.

The value of the current of a flow is obtained as a combination of the real and reactive power of a branch and the voltage value of the substation in which the branch is connected. The set of branches real power is often denoted $P$, reactive power is $Q$ and $V$ for the corresponding voltage magnitude values.

\subsection{Safety criteria and grid conduct}
TSO's Dispatchers  need to ensure that the grid operates in safe conditions at all time. For the purpose of this work, a grid will be considered safe when there are no branches in an overflowed state. In reality, a more restrictive approach is taken. TSOs often ensure that if a component of a grid were to fail (e.g. a branch, a plant or a switch) then the whole grid would remain in security \textit{i.e.} no branches are overflowed. This verification of this realistic criteria would necessitate significant additional computation resources, which is why we do not take it into account in our study.
A branch is overflowed when its flowing current is above its thermal threshold. The more current in a power line, the more it heats, which causes a dilatation phenomenon of the line. The air between the line and the ground acts as insulation and might then not been sufficient to protect nearby passengers from electric arcs. Apart from the security of passengers, a melted power line needs to be replaced. It takes several weeks to replaces lines in the context of very-hight voltage power grids and such a replacement can cost multiple million euros. Each line of the system has a nominal thermal limit, under which it is protected against melting.
By denoting $f_b$ the value of the current of the branch $b$, and $\textnormal{th}_b$ its thermal limit, an overflow corresponds to:
\begin{align}
b \textnormal{ overflowed } \iff f_b \geq \textnormal{th}_b \iff \frac{f_b}{\textnormal{th}_b} \geq 1
\end{align}

Dispatchers have a set of actions at their disposal to avoid operating the grid with failures. There are essentially three types of actions:
\begin{itemize}
\item Switching off/on branches.
\item Modifying generators electricity production.
\item Node splitting.
\end{itemize}

\textit{Branch disconnection} or reconnection is often used in nationwide power grids, although more scarcely by dispatchers. When a branch is overflowed, the system automatically disconnect the latter so that it does not melt. Dispatchers can, in theory, disconnect a branch prior to a situation of overflow for the predicted failures. Switching on/off a branch operation is cheap, because TSOs operate on element within their range (e.g. activating switches to modify the branches interconnections). In practice, dispatchers do not disconnect power lines apart for the maintenance operations, such as reparation or painting. Indeed, the power capacity of a grid only lowers when some branches are switched off (compared to a fully operating grid), which means that the grid is a priori more prone to overflows. 

\textit{Modifying the productions outputs}, often called re-dispatching, is the operation of changing the amount of energy produced by some productions. The actions is called "re-dispatching" because one production is lowered by a given amount that is redistributed among other productions. If the amount is not counter-weighted, there might not be enough productions to satisfy the demand. Re-dispatching is expensive, because this operation require the modifications of multiple generators, which are not the property of TSOs. As such, we do not take re-dispatching into account in this work.

\textit{Node splitting} represents the majority of manual interventions done by dispatchers at RTE. It is the operation of changing the interconnections configuration of elements within a substation. By definition, a substation is at the intersection of at least two branches. In fact, branches can be connected to only a subgroup of the branches connected to a substation. The operation of node splitting consists in modifying the patterns of branches interconnections. The name refers to viewing the operation as defining subnodes (or buses) to a substation, with each branch connected to none or one subnode. In the following, we limit the number of subnodes to be 2.

There are essentially two ways of operating a large-scale power grid:
\begin{itemize}
\item A preventive mode: avoiding future failures given estimations of productions and consumptions schemes
\item A curative mode: resolving a failure given the current grid loadflow
\end{itemize}
Dispatchers at RTE often ensure that the grid is currently in a secure state, i.e. with no failures. If a failure happens, dispatchers will perform curative actions. On a regular mode, without incident, they have an anticipative operating mode. Indeed, dispatchers often ensure that the "n-1" criteria is respected. This criteria means that if any element of a grid were to fail, then the grid would still be in security. Usually, they run multiple simulations of the current grid but with one of its element out-of-service, observing the resulting steady-states of the potential grids. This criteria is implemented because unexpected hazards can occur and break some element of the grid such as power lines. See Donnot's work\cite{donnotprev} on using deep learning to predict flows of a grid in "n-1" situations to accelerate those simulations.


TSOs usually operate at nation-scale. For example, RTE operates the French grid made of more than 6500 nodes, 3000 productions and 10000 branches. Taking into account hazards, maintenance, and the injections distributions, the task of operating the grid in a safe mode is rather complex. Besides, the human factor limits the tools used for predicting the grid subsequent states. For example, temperature estimations and weather predictions could be integrated when managing the grid, but would only complicate the task for dispatchers. In this context, we are interested in building a policy $\Pi: S \rightarrow A$ ($S$ is the State set and $A$ the Action set),
that would propose multiple curative solutions given grid states such that operators could take decisions rapidly by selecting an action among the candidate ones.

\subsection{Load-flow computations}

A load-flow computation is the operation to compute the values of the flows within an electric grid given the grid structure, a set of injections, and a set of parameters describing the productions, loads, branches among other elements. We make the assumption that a grid subject to injections will instantaneously converge to its-steady-state.
As such, a load-flow computation is an optimization problem subject to equations and constraints with necessitates dozens of variables for the elements describing a power grid, such as the resistance, reactance of the power lines.
\\

Most of the high voltage grids are operated in AC mode (alternative current), as opposed to DC mode (direct current). However, AC load-flows being complex and slow to compute, they are sometimes approximated with a DC approximation.

The DC modeling makes the following simplifying assumptions:
\begin{itemize}
\item Branches are lossless: the input and output powers of branches are identical.
\item All of the buses voltages are close to 1 per-unit. 
\item The branches voltage angles are almost identical. 
\end{itemize}

All these 3 assumptions make the initial problem linear as shown in the Appendix \ref{appendix:annex2}.
Making this problem linear has some advantages, including a low computation time.

On the other hand, the AC modeling does not make the previous simplifying assumptions about the system. As a consequence, AC modeling is closer to real-conditions than DC approximation. However, AC load-flow computation is harder than DC, because the overlying equations are non-linear (for example, if the difference of angles $x$ is not close to 0, then we don't have $sin(x)\approx x$, and $\sin$ is non-linear).
Multiple optimization algorithms can be used for AC load-flow computations, including Newton's method, Fast-Decoupled or Gauss-Seidel.

Our framework relies on Matpower \cite{matpower}, an open-source power system framework written in Matlab. Matpower works with the IEEE format for power grids. See Appendix \ref{appendix:annex1} for further details about this format.

\subsection{Examples of power grids}

For a first reinforcement learning challenge on power grid control, the entire French grid might be too ambitious, because we believe that the number of elements (i.e. the dimension of load-flows) is too vast for current reinforcement learning approaches (approximately 6500 nodes and nearly 10000 lines).
\\

The California power transmission grid is a relatively simpler system, which has been widely studied such as in \cite{dynamicieee}, \cite{powtrans} or \cite{ots}.
Some simplified implementations of this grid are available in open-source (and IEEE format). These implementations define various parameters about the grid elements, including the grid structure. Various version exist. We are particularly interested in the case IEEE-118, which is a simplification of the Californian grid. Explicitaly, it has 118 substations, 56 productions and 186 branches (without counting the lines between productions and substations).

\subsection{Reinforcement learning}
Reinforcement learning (RL, \cite{sutton}, \cite{rlsurvey}) is a domain of machine learning that differ from supervised and unsupervised learning. Indeed, there is no supervisor in reinforcement learning, but rather a reward signal, expressing in our case the degree of satisfaction of grid security constraints. An agent (in our case emulating a dispatcher) interacts with the Environment (in our case the game), implementing a policy determining which actions are performed given state, towards maximizing rewards. RL algorithms train a policy, which is usually a parametric function of the system state and reward so far. The set of actions in our case are taken from permitted changes in grid topology.
RL systems are often describe using Markov Decision Processes (MDP)\footnote{There exists multiple variants of MDPs. For instance, infinite MDP can involve continuous state space or action space. Partially observable MDP involve an Agent that do not directly observe the states.}. A Markov Decision Process is a tuple $\langle S, A, P, R, \gamma \rangle$ such that:
\begin{itemize}
\item $S$ is a finite set of states
\item $A$ is a finite set of actions
\item $P$ is a state transition probability matrix, \\
$P_{ss'}^a=\mathbb{P}[s_{t+1}=s'|s_t=s, a_t=a]$
\item $R$ is a reward function, \\
$R_s^a=\mathbb{E}[R_{t+1}|s_t=s, a_t=a]$
\item $\gamma$ is a discount factor $0\leq\gamma\leq 1 $
\end{itemize}

In this context, a policy $\Pi$ fully describes the behavior of an agent as a distribution of the action space given states:
$$
\Pi(a|s)=\mathbb{P}[a_t=a|s_t=s]
$$
All of the states of a MDP are Markov, i.e. the future is independent of the past given the present, which formalizes as:
$$
s_t\textnormal{ is Markov }\iff \mathbb{P}[s_{t+1}|s_t]=\mathbb{P}[s_{t+1}|s_1, ..., s_t]
$$

The \textit{return} $G_t$ of a MDP is the cumulative discounted reward starting from timestep $t$:
$$
G_t=R_{t+1}+\gamma R_{t+2}+\gamma^2R_{t+3}+...=\sum_{k=0}^{\inf}\gamma^k R_{t+1+k}
$$
From the return, we define two value functions:
\begin{itemize}
\item The \textit{state-value function} of policy $\Pi$ given state $s$ is $v_{\Pi}(s)=\mathbb{E}[G_t|s_t=s]$
\item The \textit{action-value function} of policy $\Pi$ given action $a$ applied at state $s$ is $q_{\Pi}(s, a)=\mathbb{E}[G_t|s_t=s, a_t=a]$
\end{itemize}
We further define the optimal state-value function $v*(s)$ and the optimal action-value function $q*(s, a)$ as respectively the maximum over the policies of the state-value function and the action-value function:
$$
v*(s)=\max_{\Pi}v_{\Pi}(s)
q*(s, a)=\max_{\Pi}q_{\Pi}(s, a)
$$
Optimal value functions describe the best achievable performance in the MDP, thus solving the problem. We can prove that there exist an optimal solution: its performance is greater or equal than any other model. Such an optimal policy is intrinsically optimal regarding both state-value and action-value functions.

On advantage of MDPs is their capacity of performing sequential decision making: temporal data are important in reinforcement learning (because of the delayed reward) so the data do not need to be independent and identically distributed.

\begin{figure}[h]\centering
\includegraphics[width=0.4\textwidth]{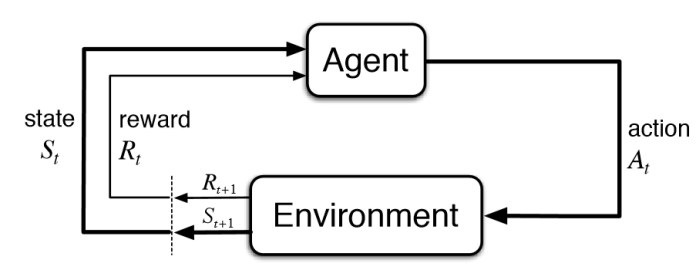}
\caption{Feedback loop in Reinforcement Learning}
\label{fig:rlmecha}
\end{figure}

The overall mechanism of reinforcement learning consists in training an Agent by interacting with an Environment. At each time step $t$, the Agent takes a set of actions $a_t$ from an Action Space, based on the current state $s_t$ of the Environment and the previous reward $R_t$. The Environment then computes the next state $s_{t+1}$ as well as a reward $R_{t+1}$ resulting from both $s_t$ and $a_t$. See Fig. \ref{fig:rlmecha} for a visual representation of this vanilla mechanism.

\section{STATE OF THE ART}
In this section, we briefly review recent works in machine learning applied to Power Systems, and introduce basic notions of reinforcement learning applicable to our problem at hand.

\subsection{Machine learning applied to Power Systems}
Reinforcement Learning is by no means the only way of applying Machine Learning to Power Systems.
There are many other ways of applying machine learning to Power Systems, not necessarily to control them, but for example to predict power flows, rather than using accurate but slow physical simulators such as Matpower or more sophisticated ones that cannot properly use GPUs.
One the advantage of machine learning is be that we do not need to explicit the laws governing the system, as it is the case for usual load-flow softwares. The models can learn inner representations of the physics concepts without any domain knowledge. 

Machine learning has been applied for Power Systems problematics. Donnot \cite{donnotprev} introduces a novel deep learning algorithm to compute load-flow, in order to replace conventional simulators based on differential equation solvers. Their approach consists in training a feed-forward multi-layer perceptron on predicting load-flows for a given grid with one line switched off. They measure the generalization of their proposed algorithm on the computations of load-flows where two lines are disconnected (which are out of the training manifold of grids with one and only one switched off line).

Their method relies on a novel architecture call Guided Dropout. It is influenced by the conventional Dropout \cite{dropout} commonly used in deep learning. However, instead of nulling randomly some hidden units, they adopt a scheme that controls the active units based on the input. More precisely, they build a feed-forward neural network with a set of productions, a set of loads, and a binary vector of line service status (switched on or off) as input. The model outputs a set of flows, that should be close to the ground-truth flows. The model is trained using back-propagation, with a regression loss such as Mean Squared Error. In this context, some hidden units are activated only when associated lines are disconnected, i.e. the associated line connectivity status input is 0. Their approach has better training and testing performance than the baseline approach that consists in the same network without Guided Dropout, i.e. all of the conditionally activated units are activated (which has more parameters, so more capacity).
\\

Neural networks have also been used for short-term load forecasting (\cite{loadfore1}, \cite{loadfore2}). The overlying idea is to better predict consumptions scheme, such that the power grid future states can be better anticipated thanks to better estimators. The recent challenge see4c (\url{http://project.see4c.eu/}) is another work on the integration of machine learning with Power Systems. The goal of the challenge was to forecast power flows given historical data.
\\

\subsection{Reinforcement learning recent works}

Here, we present some algorithms and concepts used in reinforcement learning to deal with complex systems, in the context of gaming.
\subsubsection{Deep Q-learning on Atari games}
Google DeepMind has successfully used Deep Q-learning to tackle the task of playing Atari 2600 games \cite{atarigames}, which are a collection of arcade games. The games were usually played on arcade machines, which roughly consist of a screen with a joystick used to move an object in four spatial dimensions (left/right/up/down), and one button to perform a specific action based on the type of game (e.g. shooting a laser in Space Invader). Their model achieves super-human performance on several games, and can play more than a thousand different games.
\\

To do so, their first model a Value function $Q(s, a)$ using a deep neural network, where $s$ is a state and $a$ an action. More precisely, the network is a Convolutional Neural Network \cite{lecuncnn} with ReLU \cite{relu} activations. The network input consists of 4 consecutive stacked frames (obtained from an emulator of the game), so that there is a notion of motion within the input. The network output consists in a vector of size 5 indicating which buttons are pressed.
\\

With such a modelization, an action $a*_{t+1}$ is taken, at each timestep $t$ with state $s_t$, such that it maximizes the expected Q function:
$$
a*_{t+1}=\underbrace{\textnormal{argmax}}_{a\in A} Q(s_t, a)
$$

\subsubsection{AlphaGo}
One of the most important breakthrough in the field of Artificial Intelligence over the last decade is the success of an AI over the world champion of Go. Go is a zero-sum one vs one board game, played on a board of size 19 by 19. At each non-used location, a player can put one of their piece. The goal of the game is to control the bigger area of the board. In the past, several subhuman performance models have been created, mainly based on tree search algorithms, and enhanced by trading approximation with tree depth exploration. The game of Go is significantly harder to exhaustively simulate that chess, because Go board is larger. For small $n$, there are roughly $(19\times19)^n 81^n$ reachable configurations from the current grid to a grid of $n$ more depth, which quickly falls out of scope of current computers computation capacity.
\\

In \cite{deepmindalphago}, Silver et al. present the architecture of AlphaGo. Overall, they train 3 Convolutional Neural Networks: two policy networks, and one value network. Both of the CNN take the current state of the grid as input, under the form of an image (with extra features under the scope of our discussion). The first policy network consists in copying expert moves, from an aggregation of 30 million positions, and managed approximately 57\% of accuracy on a test-set. In more details, a CNN of 13 layers with ReLU activations is trained on recorded expert moves using supervised learning. The labeled dataset consists in games of Go played by humans, which are discretized into pairs of (grid state, action taken). The output of the network consists in a grid of same size at the board (flattened for convenience). Given a set of parameters $\theta$, the network then outputs a probability distribution $p_{\theta}(a|s)$, where $a$ corresponds to every board location, and $s$ is the current state of the grid. A second lighter CNN is trained on the same task. It will be used to make rapid simulations (the authors claim $<2\mu s$). Inference is done by maximizing the networks probability function over the possible actions:
$$
a*=\underbrace{\textnormal{argmax}}_{a=(i, j)\in \{1, ..., 19\}^2}p_{\theta}(a|s)
$$

The next step of their approach consists in improving the previously learned policy by making it play against itself, using the outcome of these games as a training signal. More formally, the previous policy is trained using policy gradient learning, by making it play against previous versions of itself. Policy gradient methods are a type of reinforcement learning algorithm, which consists in optimizing parametrized policies with respect to the long-term cumulative reward using gradient descent. At this point, their trained model beat the previously best-working Go software, Pachi, in 85\% of the games.

A value network was then trained, using the third CNN, to predict the likelihood of a win, given the current game-state. This is similar to classical approach of value functions, except that it is learned in this case. They stitch the trained networks together using Monte-Carlo Tree Search. Without going into further details, AlphaGo uses a mixture of the output of the value network as well as the result of a simulation to compute the value of a state in the Monte-Carlo tree. One last trick consists in dividing the state value by the number of times a simulation has lead to this state. By doing so, there is a trade-off between exploitation (using the trained policy) and exploration (visiting new positions). The latter trick encourages exploration, since it penalizes actions that were often chosen.

\subsubsection*{AlphaGo Zero}
A more recent version, AlphaGo Zero\cite{deepmindalphagozero} achieves even better performance, not only on the game of Go, but also on Chess and Shogi\footnote{A Japanese variant of chess}. A major {\em improvement} of this version is that it does not rely on expert moves. This is an advantage because it reduces the dependence to training data (e.g. recorded games of high elo players), and leverage the importance of a simulator for reinforcement learning (in their case, the respective board games). Specifically, there is no initialization on expert behavior data. The agent learns and improves by self-playing.
\\

Apart from this improvement, the value-function neural network (the one modeling the probability of winning given a state) and the Q neural network (the one modeling the probability of an action given a state and the reward so far) are merged into a unique CNN architecture. Without going into details about the net architecture, it leverages batch normalization\cite{batchnorm} after some layers' output, on top of residual connection\cite{resnet} that improve the flow of gradient. The trained neural network is then incorporated into a MCTS algorithm to choose more consistently the investigated branches. The winner receives +1 at the end of a game, while the loser gets -1.

\section{GAME DESIGN}

This section describes the game setting that we design, such that it would lend itself to a reinforcement learning solution. Since any such game requires defining four components: {\bf State, Action, Reward, Information}, we endeavor to define first the simulation Environment and its parameterization as a State space, then the Action space and finally the Reward resulting from an observation and an action. We make explicit the Information available to the Agent to determine the next action (observable part of the State space, also called Observation space).


\subsection{Environment}
The game is based on a simulation Environment that emulates a power grid based on IEEE-118 of Matpower \cite{matpower}.
It is implemented as a Partially Observable Infinite Markov Decision Process. Formally, the environment is a tuple $\langle\,S, A, O, P, R, Z, \gamma\rangle$ such that:
\begin{itemize}
\item $S$ is a continuous set of states
\item $A$ is a finite set of actions
\item $O$ is a continuous set of observations
\item $P$ is a state transition (probability) function
\item $R$ is a reward function
\item $Z$ is an observable function
\item $\gamma$ is a discount factor
\end{itemize}


The MDP is partially observable partly because the Agents do not have access to the state resulting directly from their actions, as will become clearer in what follows. It is infinite because the productions and consumptions take real values.


We assume discrete time updates (at intervals to be determined; typically 5 minutes, 1 hour or 1 day). Actions are performed at unit time intervals. Thus, given a state $s_t$, and an action $a_t$ of the Agent, and a new set of injections ${\bf x}_{t+1}$, the variables are updated as follows by the simulation Environment:
\begin{eqnarray}
{\bf s}_{t+0.5}&=P_1({\bf s}_t, {\bf a}_t)\\ \label{eq:s}
r_{t+1}&=R({\bf s}_{t+0.5}, {\bf a}_t)\\ \label{eq:r}
{\bf s}_{t+1}&=P_2({\bf s}_{t+0.5}, {\bf x}_{t+1})\\ \label{eq:s2}
{\bf o}_{t+1}&=Z({\bf s}_{t+1}) \label{eq:o}
\end{eqnarray}
The state ${\bf s}_{t}$ includes a description of the grid topology (lines in service and line interconnections) and the status of the power flows in all lines. An action ${\bf a}_{t}$ may consist in a change in the grid topology. The reward calculation is based both on the state and the action (some states presenting more danger than others and some actions being more costly than others).

For the purpose of clarity, we decompose calculations using a half-way time step $t+0.5$ and two state transition functions $P_1$ and $P_2$. This is because the calculation of the reward $r_{t+1}$ is based on the immediate consequences of the action taken by the agent ${\bf s}_{t+0.5}$, prior to the (slower) application of a change in injections ${\bf x}_{t+1}$. In the simplest case, ${\bf x}_{t}$ can follow a defined schedule, but it could also be a random variable. Other factors may influence $P_2$, such as incidental or planned changes in grid topology.

More precisely, in Equation (\ref{eq:s}), function $P_1$ implements the laws of physics of power grid systems (it is actually deterministic in our setting). In practice, the game first applies the action onto the grid, then uses Matpower to compute the resulting flows. Equation (\ref{eq:r}) then computes the reward, depending on the last state and the actions of the Agent. Next, in Equation (\ref{eq:s2}), function $P_2$ performs another load-flow computation, based on the last state and the next set of injections ${\bf x}_{t+1}$. Finally, Equation (\ref{eq:o}) compiles the information that is made available to the agent.



The role of the Agent is to devise a strategy to make optimum actions through a policy function $\Pi(o_t; \theta)$, which may include parameters $\theta$ adjustable by training (i.e. by reinforcement learning).
The game iterates over ``LARSO'' cycles:
\begin{itemize}[\labelindent 1cm]
\item [Step L] Agent gets new observation $o_t$ and reward $r_t$ and updates/Learns (the parameters of) its policy $\Pi$.
\item  [Step A]  Agent performs Action ${\bf a}_t=\Pi({\bf o}_t)$.
\item  [Step R] Environment performs first state update (before injection change): ${\bf s}_{t+0.5}=P_1({\bf s}_t, {\bf a}_t)$, to compute  Reward: $r_{t+1}=R({\bf s}_{t+0.5}, {\bf a}_t)$.
\item  [Step S] Env. applies news injections ${\bf x}_{t+1}$ and re-computes State: ${\bf s}_{t+1}=P_2({\bf s}_{t+0.5}, {\bf x}_{t+1})$.
\item  [Step O]  Environment reveals Observation $o_{t+1}$ of $s_{t+1}$ (and reward already computed).
\end{itemize}
In this setting, two Matpower callbacks are done at steps R (with $P_1$) and step S (with $P_2$). 
The task can be parallelized for the users interested in batch reinforcement learning.

\subsection{Observation space}
An observation $o_t$ represents the state of the grid at time step $t$. Among all the variables and parameters that dictate the response of the grid to a set of injections, we keep only the changing variables of the system, detailed below. Others variables are hidden to the Agent, including the parameters of the elements constituting the grid. 

An observation is a fixed-sized structure made of the following elements:
\begin{itemize}
\item Active, reactive and voltage values of the productions
\item Active, reactive and voltage values of the consumptions
\item Active, reactive and voltage values of the lines: one 3-tuple for each substation of each line
\item Relative thermal limits
\item Lines interconnection patterns
\item Lines service status
\end{itemize}


The active, reactive and voltage values of both the productions and consumptions are the injections of the power grid at a given timestep. Each of these values are stored as lists of fixed sized throughout the game. For IEEE-118, the lists are of size 56 for productions, and of size 99 for consumptions.
\\

The line power flow values are stored similarly. We keep two values per line: the in-flow and the out-flow. This is justified by the fact that there are losses within lines in the AC setting. For IEEE-118, there are 186 lines.
\\

The relative thermal limit vector is the element-wise division of the lines flowing current list by the lines thermal limits list. More precisely, given a set of flowing currents ${\bf f}_t=({\bf f}_{i, t})_i$ and a set of associated thermal limits (one per branch) $T=({\bf th}_i)_i$ (thermal limits are fixed through time), the relative thermal limit $({\bf r}_{i, t})_i$ is:
\begin{center}
$\forall i, {\bf r}_{i, t} = \displaystyle\frac{{\bf f}_{i, t}}{{\bf th}_i}$
\end{center}
The modelization condenses both the values of the current and the values of the thermal limits for every line. Consequently, a line $i$ is overflowed iff ${\bf r}_i \geq 1$.
\\

The information about the grid topology is given using a \textit{topology list} noted $\tau$ of fixed dimension. Each element $i$ of the topology list represents the id of the chosen configuration for the substation $i$, except that the id are converted to one-hot vectors. Given a substation $i$ with $n_i$ elements, and our hypothesis of a maximum of two buses per substation, the following formula gives the number of (non-unique) possible topological configurations with all objects considered switched on\footnote{With two possible groups of directly connected elements, and no elements disconnected (i.e. each element is directly connected to one group), then choosing one configuration is equivalent to choosing the number of elements $k$ to pick for one group, and then choosing $k$ elements. Choosing $k$ elements among $n$ is $\binom{n}{k}$, which is summed to take into account all the possible number of elements in a group.}:
$$
\sum_{k=0; k\neq 1, k\neq n-1}^{n} \binom{n}{k}=2^{n}-2n
$$
For identifiability issues, since we consider configurations to be equal up to bus permutation, there are $2^{n-1}-n$ unique possible configurations for a substations of $n$ elements (since we counter twice each configuration in the previous formula). We use a one-hot encoding of such configurations, i.e. if a substation (i) has $n_i$ configurations, we will use a sub-vector of dimension $n_i$ with a 1 in the $j^{th}$ position if the $j^{th}$ configuration is used and 0 everywhere else.

\begin{figure}[ht!]
\centering
\includegraphics[width=\linewidth]{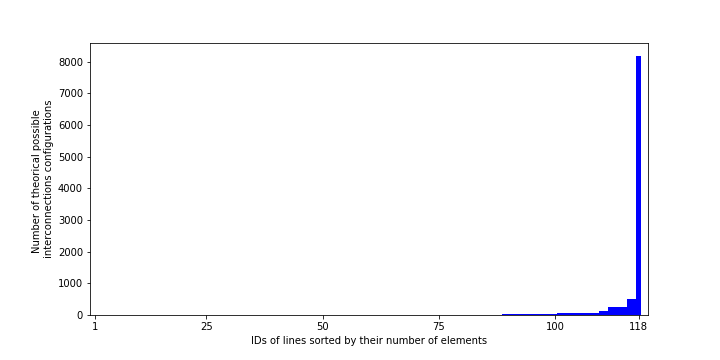} 
\caption{Distribution of the number of configurations per substation in IEEE-118. The bar graph was obtained by sorting the substations by their number of possible configurations, without restrictions. We can infer that only one line is responsible for the majority of the sum of number of configurations of the whole grid. In fact, most substations of the IEEE-118 have at most 10 lines interconnection configurations.}
\label{fig:distribconfig}
\end{figure}

For IEEE-118, this approach leads to a topological list of sum of its elements size approximately 10000. However, we note that only one substation is responsible for three quarters of the size, because there are 14 objects connected to it. See Fig. \ref{fig:distribconfig} for a plot of the distribution of the total number of (unique) interconnection configurations per substation, with all elements non-disconnected. Consequently, in order to limit the size of the topological space, we reduce the number of available topologies for some of the bigger substations.\\

Note that our design of the topological modelization of the grid involve the assumption that all of the objects are connected. The number of all the possible configurations of a substation, where some elements can be disconnected is
$$
\sum_{l=0}^n\sum_{k=0; k\neq 1, k\neq l-1}^{l} \binom{l}{k}=2\sum_{l=0}2^{l-1}-l=O(2^n)
$$
We propose to decouple the representation of the topology into two parts: a vector of line status and a one-hot vector representing the configuration as if all the lines were in-service. The lines service status vector is of size the number of lines in the grid and takes binary values: 0 represents a line out-of-service, 1 a line in-service.
There are exactly $2^{n-1}-n+n=2^{n-1}$ values to fully represent this modelization instead of the $O(2^n)$ when considering configurations with out-of-service elements. 



\subsection{Action space}

The game allows two types of actions: the disconnection and reconnection of lines, and the modification of the grid topology. For better integration within the Gym environment, those two types of action are stored within an Action 2-tuple such that the players need to provide one structure at each time step. Besides, the game validates that actions proposed by the player are well formed, as described below. 
\\

\subsubsection{Changing the line service status}

The line service status is encoded as a vector $({\bf a}^1_{i, t})_i$ of size the number of lines of the grid (186 for the IEEE-118) such that:

\begin{equation*}
  \forall i, {\bf a}^1_{i, t} =
    \begin{cases}
      1 & \text{: switch line $i$ on}\\
      -1 & \text{: put line $i$ out-of-service}\\
      0 & \text{: do nothing to line $i$}
    \end{cases}       
\end{equation*}

Given a line service status ${\bf s}^{\textnormal{line status}}_{i, t-1}$ from the previous observation, and a line service status action ${\bf a}^1_{i, t}$, the observation is updated as follow:

\begin{equation*}
  \forall i, {\bf s}^{\textnormal{line status}}_{i, t} =
    \begin{cases}
      {\bf a}^1_{i, t} & \text{if }{\bf a}^1_{i, t}\neq 0\\
      {\bf s}^{\textnormal{line status}}_{i, t-1} & \text{otherwise}
    \end{cases}       
\end{equation*}


The verification step for an action consists in checking that the lines service action is of the expected size and that the values are -1, 0 or 1.
\\

\subsubsection{Changing the grid topology}


The grid topology action is encoded as a vector $(\tau_i, t)_{1\leq i \leq n_{\textnormal{substations}}}$ of expected size the number of substations in the grid (118 for IEEE-118). Its values $\tau_{i, t}$ are either None or a one-hot vector whose active value indicate the id of the associated substation target topological configuration. In other words, players or Agents specify, for each substation, whether to not change the local topology (value of None) or to update the local topological configuration of the substation to the desired configuration.
\\

One approach for the topological vector would have been to specify, for each substation and for each of its connected object, the node on which the object is connected. However, we find that there are identifiability issues with such an approach, which could prevent reinforcement learning models from learning good representations for the grid.
\\

The verification step for the topological action is assert that the latter is of expected size, and that its elements are either None, or a one-hot vector of expected shape.
\\

\subsection{Reward}

The reward is designed as a sum of 4 subrewards, each intended to focus on one aspect of grid conduct:
\begin{enumerate}
\item Line usage subreward
\item Cut load subreward
\item Action cost subreward
\item Distance to the reference grid subreward
\end{enumerate}
In the following, we define and give insight about each subreward.
\\

\subsubsection*{Line usage subreward}
Ideally, dispatchers should avoid situations where a line is overflowed. Given a timestep $t$, we note $(fa_{i, t})_i$ the set of current flows and $(th_i)_i$ the set of thermal limits. We can use the following formula to count the number $n_o$ of overflowed lines:
\begin{align*}
n_o = \sum_{i=1}^{N_{\text{lines}}}\floor*{\frac{fa_{i, t}}{th_i}}
\end{align*}
However, this modelization does not give any information on the usage of the lines, except for the overflowed ones. With such a formula, there are no explicit way to discriminate two situations where the number of overflowed lines are identical, since the reward would be identical. Ideally, we would like all the lines to use as little of their capacity as possible, rather than some lines exploding their limits and others using close to nothing. Besides, another drawback of the formula can appear when some lines have a ratio slightly below 1, and others slightly above 1. The first group would not be considered as overflowed while the second would increase the reward. In other words, this formula is highly sensitive to noise when the ratio are close to 1. Because of the flaws, we introduce a modified formula that we call the line usage reward:
\begin{align*}
\sum_{i=1}^{N_{\text{lines}}}\big(\frac{fa_{i, t}}{th_i}\big)^2
\end{align*}
Note that we use the squared of the ratio for computing the line usage. This allows to have non-negative ratio, and also to amplify the impact of overflowed lines and minimize the impact of secured lines. Challengers can modify the reward by using an absolute value instead of the square. The subreward is multiplied by -1 before being sent to the players, such that models will minimize line usage.
\\

\subsubsection*{Cut load subreward}

A major aspect of grid conduct is to carry electricity such that every consumption has the expected active and reactive values. If a consumption is cut, this means that a group of people won't have access to electricity for a certain amount of time. We would like to avoid these situations at all cost. Consequently, by design, the game will stop the current playing epoch once a load has been cut, i.e. there are not enough incoming electricity to satisfy the local demand. When such an event happens, the game will return a specific reward; it is up to the player to load the next epoch.

In real life, situations might happen when the grid has overflowed lines. In that case, those lines are dynamically disconnected to protect them. After such disconnections, the grid will naturally converge to an equilibrium consequently to the topology and the laws of physics. When the equilibrium is reached, other lines can then be overflowed, since the whole grid has the same injections but a lower capacity. Recursively, this can create a \textit{cascading failure}, where disconnected overflowed lines provoke new overflowed lines, which could eventually isolate a consumption. The game consequently has a cascading failure module that will simulate cascading failures after an Agent has taken an action. If the cascading failure does not disconnect any consumption, the reward remains unchanged. On the other hand, if a load is disconnected, then the game will stop the epoch and retrieve the corresponding load cut reward.
\\

\subsubsection*{Action cost subreward}

The cost of putting a line out-of-service or changing the topology (pattern of line interconnections) is integrated within the reward computation. It is motivated by real-life conditions, where those actions need to be performed manually by specialized teams and at specific locations. The costs of one line disconnection, one line reconnection, or one substation topological change are identical. The action cost reward sums the cost of those atomic operations for every action taken by the Agent to better. More precisely, the value of this sub-reward is the cost one one action, multiplied by the number of disconnections added to the number of re-connections and the number of topological changes.
\\

\subsubsection*{Distance to the reference grid subreward}

Another aspect of grid conduct gravitates around the idea that dispatchers perform well with a given topological setting. We would like the Agents to ultimately change the topology of the grid in response to potential harms, such that the grid topology is not far from a reference topology. This subreward computes the distance of a grid to a reference grid by summing the number of local topological changes to transform the former into the latter.

\section{RESULTS}
In order to demonstrate use cases of the proposed environment, we developed basic baselines relying on hand-crafter algorithms. We measured the performance of each model by running similar experiments. Besides, we applied models to resolve crisis situations, where a power grid suffers from line power overflow to be eliminated.

\subsection{Baselines implementations}
\subsubsection{Do-nothing policy}
The Agent does not take any action.

\subsubsection{Random line-disconnection policy}
The grid has one and only one disconnected branch (line put out-of-service) at each time step, chosen randomly by the Agent. Equivalently, the Agent choses a random branch to disconnect and reconnects the previous disconnected line.

\subsubsection{Random node splitting policy}
The Agent selects one substation at each time step, and randomly changes its local topological configuration. Note that topology changes are not reverted apart from the action of the Agent: they perpetuate in time until further changed.

\subsubsection{Greedy line-disconnection policy}
At each time step, the policy simulates every 1-line disconnections and applies the action that maximizes the reward. Formally, the Action Space $A$ is of size 186 for the IEEE-118 and is made of every branch disconnection possible. This is equivalent to a Tree Search of root the current state $s_t$, with leaves $i$ being the reward for disconnection of branch $i$ and 1-line disconnection as actions. The policy choses the optimal action 

$$
a_t^{*}=\underbrace{\textnormal{argmax}}_{a_t\in A}R(s_{t+1}, a_t)
$$

We are in the process of obtaining benchmarks on these baselines, using a same-context experiment. 



\subsection{A practical case: curing a crisis situation}
\begin{figure}[h]\centering
\includegraphics[width=0.35\textwidth]{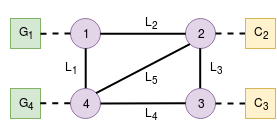}
\caption{Illustration of case4gs. The purple circles are the 4 substations numbered from 1 to 4, the 2 productions are the green squares $G_1$ and $G_4$, and the 2 consumptions are the yellow ones $C_2$ and $C_3$. This is the reference grid for our experiment, and no elements are added throughout the game (only disconnections, and substations inner lines interconnections).}
\label{fig:case4gs}
\end{figure}

We conduct an experiment treating a toy use case in the proposed framework. The grid used in this experiment, displayed in Fig. \ref{fig:case4gs}, is made of only 4 substations, 2 productions, 2 consumptions and 4 branches. By construction, we set every branch to a thermal limit threshold of 100MW.

With this example grid, we devised a toy experiment consisting in playing only two time steps of the game, with hand-selected injections detailed below and in DC approximation (such that we can neglect reactive values). The Agent is the Do-Nothing Agent, that do no take any action throughout the whole game. The injection will create an overflow, and we show a curative action that induces a higher reward. Specifically, with the notations of Fig. \ref{fig:case4gs}, the injections are displayed in Fig. \ref{fig:injectionstable}.
\begin{figure}
\centering
\begin{tabular}{ |p{0.11\textwidth}||p{0.04\textwidth}|p{0.04\textwidth}||p{0.04\textwidth}|p{0.04\textwidth}| p{0.04\textwidth}|p{0.04\textwidth}|  }
 \hline
 & $G_1$ & $G_4$ &  $C_2$ & $C_3$\\
 \hline
 Time step 0& 150 & 50 & 50 & 150 \\
 \hline
 Time step 1& 200 & 50 &  100  & 150\\
 \hline
\end{tabular}
\caption{Precomputed values of the injections to be loaded at each time step of the game.}
\label{fig:injectionstable}
\end{figure}
Note that in for each time step, the sum of production equals the sum of consumptions. This is induced by the DC approximation for which the lines are lossless.



\begin{figure}[ht!]
\centering
\begin{subfigure}{.4\textwidth}
  \centering
  \includegraphics[width=\linewidth]{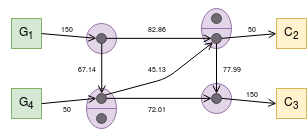}
  \caption{Steps L and A: time step $t=0$, initial grid configuration (observation $o_0$). The Do-Nothing Agent agent performs no learning and no action.}
\end{subfigure}
\\
\begin{subfigure}{.4\textwidth}
  \centering
  \includegraphics[width=\linewidth]%
  {case4timestep0.png}
  \caption{Step R: state $s_{0.5}$ updated from $s_0$ after the Do-Nothing Agent has made his action (none); the resulting load-flow state is the same as the initial state, in (a). The associated reward given $s_{0.5}$ and $a_0$, $r_0$ is -2.468.}
\end{subfigure}
\\
\begin{subfigure}{.4\textwidth}
  \centering
  \includegraphics[width=\linewidth]%
  {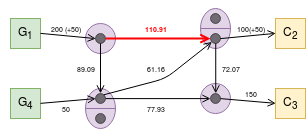}
  \caption{Step S and step O: observation $o_1$ of the grid, obtained by applying the new injections $x_1$ of Fig. \ref{fig:injectionstable}, then computing a load-flow, and exporting a view of $s_1$. Note that $o_1$ contains one overflowed line: the Agent should find an action supposed to cure the overflow, or the situation could lead to a global outage.}
\end{subfigure}
  \caption{Explicit representations of the evolution of a grid for the time step $t=0$. The purple ellipses and circles are the substations, the yellow squares are consumptions and the green squares are productions. The values written next to each line represents the real power flowing into the associated line. The direction of the arrows indicate the direction of the flowing current in the line.}
  \label{fig:caseanalysis}
\end{figure}

Between time step 0 and time step 1, the consumption $C_2$ rises by 50MW. Generator $G_1$ is incremented by the same amount to produce enough electricity. A representation of the initial state of the game, i.e. the first state observed by the do-nothing Agent, is displayed in Fig. \ref{fig:caseanalysis}(a). The substations 2 (top right) and 4 (bottom left) have two nodes. This comes from the fact that they are made of four elements (substation 2 has three power lines and one consumption, substation 4 has three power lines and one generator). On the contrary, substation 1 (top left) and 3 (bottom right) only have one node, because there cannot be more than one group of directly connected elements, such that there is at least two elements per group (because electricity need to exit). For both the two-nodes substations, the initial lines interconnection configuration is to have all elements directly connected, i.e. on the same bus.

For step A, the do-nothing policy will not output any action for the timestep 0. Formally, the Environment will apply the action of the player onto the grid, and discard the flows that are not pertinent anymore (because the flows are a function of the injections, and the grid explicit topology). This subsequent grid, obtained by computing a load-flow using Matpower, after taking the previous set of injections and performing no topological change is the same as Fig. \ref{fig:caseanalysis}(a). At this point, there are no overflow, so no cascading failure simulation, so the subreward of cut load, $r_{\textnormal{load cut}}=0$. With the hypothesis that the the initial grid is the reference grid, the subreward of the distance to the reference grid, $r_{\textnormal{distance to ref}}=0$. No action were performed, so the cost of the operation, $r_{\textnormal{action cost}}=0$. Finally, we compute the subreward $r_{\textnormal{line usage}}$ of the line capacity usage is
\begin{align*}
r_{\textnormal{line usage}}&=-\sum_{l=1}^{l=5} \big(\frac{f_i}{\textnormal{th}_i}\big)^2=-\frac{1}{100^2}\sum_{l=1}^{l=5} f_i^2\\
&=-\frac{82.86^2+67.14^2+45.13^2+77.99^2+72.01^2}{100^2}\\
r_{\textnormal{line usage}}&\approx -2.468
\end{align*}
because by construction of this situation, all of the thermal limits are 1. Note that the flows of the lines connecting productions and consumptions to the rest of the system do not count in the reward computation, since there values are do not depend on the Agent. The timestep computed reward is then:
\begin{align*}
r_0&=r_{\textnormal{load cut}}+r_{\textnormal{distance to ref}}+r_{\textnormal{action cost}}+r_{\textnormal{line usage}}\\
r_0&=-2.468
\end{align*}

The Environment returns the previous reward to the Agent, and then will compute the state ${\bf s}_1$ of the Environment: this is step S. To do so, the game will load the injections $x_1$ of the timestep 1, displayed in Fig. \ref{fig:injectionstable}, and compute a load-flow to retrieve the subsequent flows. The resulting state ${\bf s}_1$ is detailed in Fig. \ref{fig:caseanalysis}(c). As depicted by the thick red line, one line is overflowed. At this point, the game has done one full cycle of LARSO.

\begin{figure}[ht!]
\centering
\begin{subfigure}{.4\textwidth}
  \centering
  \includegraphics[width=\linewidth]{case4timestep1.png}
  \caption{Steps L and A: time step $t=1$, previous grid configuration (observation $o_1$) from the previous time step (see Fig. \ref{fig:caseanalysis}. The Do-Nothing Agent agent performs no learning and no action.}
\end{subfigure}
%
\begin{subfigure}{.4\textwidth}
  \centering
  \includegraphics[width=\linewidth]{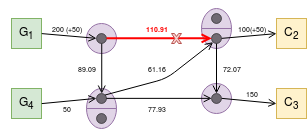}\\
  \includegraphics[width=\linewidth]{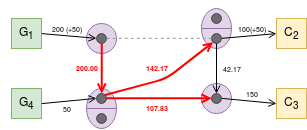}\\
  \includegraphics[width=\linewidth]{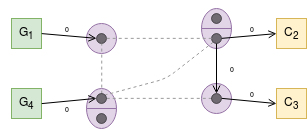}
  \\
  \caption{Step R: Cascading failure simulation. The action of the Agent leads to a situation where at least one line is overflowed (top line). For computing the hidden state $s_{1.5}$, the Environment will perform a cascading failure simulation, which consists in successively and repeatedly switch off overflowed lines, and computing a new load-flow. This is part of the function $P_1$ of step R. First row= cascading simulation initialization; second row= top line put out-of-service resulting in 3 more over-flowed lines after a load-flow computation; third row=new over-flowed lines out-of-service leads to global outage. In that case, the reward $r_1$ returned by the Environment is equal to the value of the subreward related to a cut load.}
\end{subfigure}%
\caption{Steps L, A and R of timestep $t=1$. In that case, the cascading failure simulation provoked a global power outage, which leads to a game over.}
\label{fig:cascfailure}
\end{figure}

Upon such state, the do-nothing policy does not perform any action: step L and A for timestep $t=1$. To compute $s_{1.5}$, since there is an overflowed branch, the game will simulate a cascading failure: repeatedly making overflowed lines out-of-service, then computing the consequent load-flow. This leads to the successive rows in Fig. \ref{fig:cascfailure}(b). The first row results from disconnected the overflow line; the second row is the consequent load-flow, which provokes 3 additional overflows; the last row displays the next line disconnection, which underlines a global outage of the power system.

The outage induce that at least one consumptions has been cut. This is a situation of game over: the game will reinitialize the grid overall structure, and load the the remaining timesteps to be played. Before that the Environment returns the value of the cut load subreward.
\\

\begin{figure}
\centering
\begin{subfigure}{.4\textwidth}
  \centering
  \includegraphics[width=\linewidth]{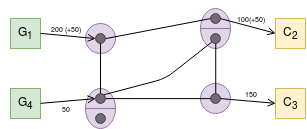}
  \caption{Step A: an Agent applies a node-splitting aciton on the top right substation, given the observation $o_1$ of Fig. \ref{fig:caseanalysis}(c). Note that we do not display the flows here (they first need to be computed by step R below).}
\end{subfigure}%
\\
\begin{subfigure}{.4\textwidth}
  \centering
  \includegraphics[width=\linewidth]{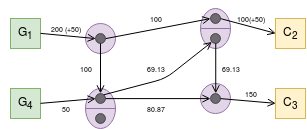}
  \caption{Step R: state $s_{1.5}$ obtained after computing the load-flow of the above state. Contrary to the do-nothing policy, there are no overflow here, thus no power outage (and also no cascading failure simulation). We would expect a higher reward than in the case of outage (depending on the load cut subreward value).}
\end{subfigure}%
\caption{A candidate action that can avoid the crisis situation of outage, given observation $o_1$. The Agent applied a curative node-splitting action, which does not provoke any overflow once applied. In that case, the Agent do not lose the game, and subsequent following LARSO cycles can be performed.}
\label{fig:case4alternative}
\end{figure}

We show in Fig. \ref{fig:case4alternative} a node splitting action that can avoid the outage given the previous state $s_1$. Specifically, we directly connect the top right consumption with the top left substation, and the bottom right substation with the top right one.

\section{CONCLUSION AND FUTURE WORK}
In this work, we try to tackle the task of applying machine learning, and more specifically reinforcement learning, on the task of operating a extra high voltage power grid in safe conditions. To do so, we propose a novel game environment that is able to simulate a power grid through time given an Agent policy. For reinforcement learning integration, the game is modeled using a Partially Observable Infinite Markov Decision Process. At each time step, an Agent takes an action given an observation, which leads to a first inner state that is used to compute a reward (along with the chosen action). The game then loads the next set of injections, compute the resulting state and provides the Agent with a new observation. Our approach can be used with multiple level of simplicity, notably by the type of algorithm to be used to compute load-flows. States are not directly visible by the Agents, but rather exported into observations. The reward computation involves multiple inner subrewards, which are designed to reflect some of the goal of TSOs dispatchers. We demonstrated how the game processed two timesteps, with a do-nothing Agent and compare it with an Agent that performs one grid modification.
\\

For future work, we plan to design game situations that are interesting for the player and include situations of crisis leading to eventually cascading effects, i.e. meta-stable states, that pose difficult problems to be solved. We need to reverse-engineer the problem in an adversarial way in some sense: one player creates difficult problems and the other tries to solve them. The game will come with a Graphical User Interface, which will allow users to manually play the game, and could be used to watch policies in action.
integrate new features to the game in order to make it closer to real-life conditions. 
Maintenance schemes and random branches hazards will be furtherly integrated, along with random noise injection into the planned productions and demands. Once the user interface is operational, we plan to spend time optimizing the code such that computing can be fast enough to tackle the number of necessary learning steps.
We will also establish more complex baselines, focusing on additional approaches for reinforcement learning including Actor-Critic methods (\cite{actorcritic1}, \cite{actorcritic2}) and Normalized Advantage Functions (\cite{naf}).
Those baselines performance will help use tune a set of suitable hyperparameters, such as subreward values. Besides, new grid parameters will be explored for the IEEE-118, including thermal limits, such that we can tune the game difficulty.

\vspace{1.5cm}
\subsubsection*{Acknowledgements}
This work was supported by INRIA and ChaLearn. Special thanks to my collaborator Kimang Khun and my advisors Isabelle Guyon, Antoine Marot, Benjamin Donnot. I am grateful to Marc Schoenauer for welcoming me at the LRI lab in the TAU group and to RTE's advisors including Patrick Panciatici for their guidance.

\bibliographystyle{unsrt}

\newpage
\onecolumn

\begin{appendices}

\section{About the IEEE format for power grids}
\label{appendix:annex1}
The IEEE format allows to represent a steady-state of a power grid in a condensed manner. We use the version 2 of this format, which differs from its first iteration by the variables it stores. IEEE format is used to compute load-flows using the open-source software Matpower.
\\

More precisely, the IEEE format is made of at least three matrices (+ one version value and one value indicating the base MVA of the system):
\begin{itemize}
\item The bus matrix: stores values about the substations and the consumptions
\item The generator matrix: parameters related to the generators
\item The branch matrix: stores the values related to the flows of the system
\end{itemize}

Here, we list the parameters stored in each of those three matrices. For each one, the columns indicate parameters, and the lines indicate the corresponding objects.
\subsection*{Bus matrix}
The bus matrix is a matrix of shape $n\times 13$, which implies that there are 13 parameters for every bus. The parameters are (i.e. the columns):
\begin{enumerate}
\item ID of the bus
\item Type of the bus: 1 for a PV bus, 2 for a PQ bus, 3 for the slack bus (or reference bus), and 4 for isolated bus (not linked to any other element)
\item Real power demand
\item Reactive power demand
\item Shunt conductance (some substations have shunts)
\item Shunt susceptance
\item ID indicating the area of the bus (not used in our case)
\item Voltage magnitude
\item Voltage angle
\item Base voltage (total voltage is magnitude times base voltage)
\item ID indicating the zone of the bus (bot used in our case)
\item Maximum voltage magnitude
\item Minimum voltage magnitude
\end{enumerate}

Some of these parameters, such as the voltage magnitude, need to be specified in per-unit. It is the expression of some quantities as fractions of a defined base unit quantity (for voltage magnitude, the base unit quantity is the base voltage parameter). We do not explicitly use the area of zone parameters, since we consider every elements to be within the same grid.
\\

One thing to note about the IEEE format is that there are no notion of substations. In fact, matpower only used the notion of bus. We use some tricks, which include artificially created nodes with the same parameters, for actions such as node splitting.

\subsection*{Gen matrix}
The matrix of generators has $n$ lines, where $n$ is the number of generators of the grid, and 10 parameters, which are:
\begin{enumerate}
\item ID of the bus directly on which the generator is directly connected
\item Real power output
\item Reactive power output
\item Maximum reactive power output
\item Minimum reactive power output
\item Voltage magnitude setpoint
\item Base MVA of the generator
\item Status of the generator (0 out-of-service, $>$0 in service)
\item Maximum real power output
\item Minimum real power output
\end{enumerate}

\subsection*{Branch matrix}
The branch matrix has 11 parameters, and 4 extra values per branch representing the flows. Branch are identified using a {\em from} bus and a {\em to} bus (used for convention). The 11 parameters for a branch are:
\begin{enumerate}
\item ID of from bus
\item ID of to bus
\item Resistance
\item Reactance
\item Susceptance
\item Long term rating
\item Short term rating
\item Emergency rating
\item Transfromer shift phase angle
\item Branch status (1 in service, 0 out-of-service)
\end{enumerate}

On top of that for steady-states, there are 4 extra columns:
\begin{enumerate}
\item P at origin
\item Q at origin
\item P at destination
\item Q at destination
\end{enumerate}

Branch are represented using origin and destination values, since in the AC mode there can be losses (function of some parameters including the branch resistance).

\newpage
\section{Power Flows equations}
\label{appendix:annex2}
This section is a rapid overview of the problem that power flow software need to solve.

\subsection{Model of the power grid}
Let G be a grid with n nodes, m power lines. \\
The nodes of G are divided in two parts, namely the \textit{generator nodes}, where at least one production unit (power plant, wind plant etc.) participating to voltage control is connected\footnote{Actually, for the system to be properly specified, one node where there is a generator will be called a \textit{slack bus}}, and those called \textit{load nodes}\\. 

To connect node $i$ and node $j$ there are element with complex impedance $Z_{i,j}$. If nothing connects the two, one can think of  $Z_{i,j} = \infty$. \\
Often, it is more convenient to think of the admittance $Y$, instead of the impedance $Z$. The admittance is nothing more than :
\[ Y_{i,j} = \frac{1}{Z_{i,j}}\]
\noindent So if two nodes $i$ and $j$ are not connected, we have $Y_{i,j}=0$. \\

The Ohm's law (also called Kirchoff's voltage law) between node $k$ and node $j$, in complex form can be written as :
\[i_{k \to j} = Y_{i,j}\times(V_j - V_k) \]

There is another fundamental law in a power grid, the Kirchoff's power law. It states that, at a node $i$ :
\[ i_{k} = \sum_{j=1, j\neq i}^{n_{\text{node}}} i_{k \to j}\]
\noindent where $i_{k}$ is all the complex current injected at node $k$ and $\forall k, i_{k \to j}$ denote the (complex) current flowing from node $k$ to node $j$. \\

If we denote by $Y$ the matrix :
\[ Y =  \left[
\begin{array}{cccc}
\sum_{j \neq 1} Y_{1,j} & -Y_{2} & \dots & -Y_{n} \\
-Y_{1} & \sum_{j \neq 2} Y_{2,j} & Y_{2,3} & \dots \\
\vdots & & & \\
Y_{1,n} & Y_{2,n} & \dots & \sum_{j \neq n} Y_{2,j}
\end{array} \right] \]
\noindent and  substituting  Kirchoff's voltage law in Kirchoff's current law we have :
\[\left[ \begin{array}{c}
i_1 \\
\vdots \\
i_n
\end{array} \right] = Y
\left[ 
\begin{array}{c}
v_1 \\
\vdots \\
v_n
\end{array} \right]
\]

$Y$ is commonly called the admittance matrix. \\

\subsection{Equations to satisfy}
A load-flow is a computation that takes as input:
\begin{itemize}
\item the real power for all load nodes $P_D$
\item the reactive power for all loads nodes $Q_D$
\item the real power for all generator nodes $P_G$
\item the voltage magnitude $|V|$ for all generator nodes
\item the voltage angle $\Theta$ for the slack bus
\item the voltage magnitude $|V|$ for the slack bus
\end{itemize}
With these informations, a load-flow computes, for each load-bus the voltage angle $\Theta_l$ and magnitude $|V|_l$ and then derived other the interesting quantities, such as the active power flow, the reactive power flow, or the currents power flow on each power line of the system.\\

The power flow equations are, for each node (slack node, production node or load node) $i$ of the power grid:
\begin{align*}
0 &~= - P_i + \sum_{k=1}^N |V|_i|V|_k \left( G_{i,k}.\cos(\Theta_i - \Theta_k) + B_{i,k}\sin(\Theta_i - \Theta_k) \right) & \text{ for the real power} \\
0 &~= Q_i + \sum_{k=1}^N |V|_i|V|_k \left( G_{i,k}.\sin(\Theta_i - \Theta_k) - B_{i,k} \cos(\Theta_i - \Theta_k) \right) & \text{ for the reactive power}
\end{align*}
\noindent where:
\begin{itemize}
\item $P_i$ is the real production injected at this node
\item $G_{i,k}$ is the real part of the element in the bus admittance matrix, eg the real part of the admittance of the line connecting bug $i$ to bus $k$ (if any) or $0$ (if not) 
\item $B_{i,k}$ is the imaginary part of the element in the bus admittance matrix, eg the imaginary part of the admittance of the line connecting bug $i$ to bus $k$ (if any) or $0$ (if not)
\end{itemize}

For the system to be fully determined by these sets of equations, these equations are not written for the slack bus, and only the real part of this equation is written for the production nodes. \\

Once these quantities have been computed, one can compute the active power flows on each elements of the network. For example, for a given line connecting bus $i$ to bus $k$ with admittance Y at the origin node $i$ having conductance $S_i$ and susceptance $B_i$:
\begin{align}
P_{i\to k} &=~ |V_i|.|V_k|*Y.\sin(\Theta_i - \Theta_k) + |V_i|^2.S_i\\
Q_{i\to k} &=~ -|V_i|.|V_k|*Y.\cos(\Theta_i - \Theta_k) + |V_i|^2.\left( Y - B_i\right) \label{eq:ac_q} \\
I_{i\to k} &=~ \dfrac{\sqrt{P_{i\to k}^2+Q_{i\to k}^2}}{|V_i|} \label{eq:ac_i}
\end{align}

\subsection{DC approximation}
For a more detailed information, the powerflows are shown in \href{http://home.eng.iastate.edu/~jdm/ee553/DCPowerFlowEquations.pdf}{DCPowerFlowEquations.pdf}. This section is greatly inspired from\\
\href{https://www.mech.kuleuven.be/en/tme/research/energy_environment/Pdf/wpen2014-12.pdf}{DC power  ow in unit commitment models} chapter 3. In this section we will suppose that there is not transformers nor phase shifters. These two objects can of course be taken into account in the DC approximation, as shown in the two previous papers.

In this part, we will present one of the most used model to approximate the load-flow equations. In counterpart, some results of the AC model won't be accessible for example the losses or the voltage magnitudes. \\
Despite these drawbacks, DC modelisation has two main advantages. First of all, it can always find a solution to its equations, and more importantly it is much faster to compute.\\

Let's recall the powerflow equations in the AC case:
\begin{align*}
0 &~= - P_i + \sum_{k=1}^N |V|_i|V|_k \left( G_{i,k}.\cos(\Theta_i - \Theta_k) + B_{i,k}\sin(\Theta_i - \Theta_k) \right)
& \text{ for the real power} \\
0 &~= Q_i + \sum_{k=1}^N |V|_i|V|_k \left( G_{i,k}.\sin(\Theta_i - \Theta_k) - B_{i,k} \cos(\Theta_i - \Theta_k) \right) 
& \text{ for the reactive power} \\
\end{align*}

The DC modeling will make three important assumptions:
\begin{enumerate}
\item the resistance (R) of a line is negligible its reactance (X)
\item For two connected buses (let's say $i$ and $k$) the difference of phase $\Theta_i - \Theta_k$ is very small
\item The voltage magnitude at each bus is equal to its nominative value. \\
\end{enumerate}
\noindent The impact of each of these assumptions on the power-flow equation are discussed now:

\begin{enumerate}
    \item R $<<$ X. The part has a big impact on the equations. First this induces that the losses are fully neglected.\\

For every line, the admittance can be written:
\begin{align*}
Y &= \frac{1}{Z} 
= \frac{1}{R + jX} 
 = \frac{R^2}{R^2+X^2} - j\frac{X}{R^2+X^2}
\end{align*}
And by definition, we have :
\[ Y = G + jB\]
\noindent thus : 
\begin{align*}
G &= \frac{R^2}{R^2+X^2} 
\underset{R \to 0}{\to} 0 \\
&\text{and} \\
B &= \frac{-X}{R^2+X^2} \underset{R \to 0}{\to} \frac{-1}{X} \\
\end{align*}

\noindent So the power flow equations become:
\begin{align*}
0 &~= - P_i + \sum_{k=1}^N |V|_i|V|_k \left(B_{i,k}\sin(\Theta_i - \Theta_k) \right) &\text{ for the real power} \\
0 &~= Q_i + \sum_{k=1}^N |V|_i|V|_k \left(B_{i,k} \cos(\Theta_i - \Theta_k) \right) &\text{ for the reactive power} \\
\end{align*}

\item $\Theta_i - \Theta_k \approx 0$. This will allow a linearization of the problem, as the trigonometric functions $\sin$ and $\cos$ will be approximate by the identity and the constant $1$ (first order approximation). The powerflow equations then becomes:
\begin{align*}
0 &~= - P_i + \sum_{k=1}^N |V|_i|V|_k B_{i,k}(\Theta_i - \Theta_k ) 
& \text{ for the real power} \\
0 &~= Q_i + \sum_{k=1}^N |V|_i|V|_k B_{i,k}  
& \text{ for the reactive power} \\
\end{align*}

\item $|V|_j \approx |V|_{\text{nom}}$. The last non linearity in the previous equations arises due to the factor $|V|_i.|V|_k$. Assuming that $|V|_j \approx |V|_{\text{nom}} $ will make them disappear. This is also a very strong assumption preventing us from getting voltage magnitude as a results of the DC approximation. This leads to:
\[ |V|_i|V|_k \approx |V|_{\text{nom}}*|V|_{\text{nom}}\]

So at the end the equations are:
\begin{align}
P_i &~= \sum_{k=1, k \neq i}^N  B_{i,k}(\Theta_i - \Theta_k ) & \text{ for the real power} \label{eq:dc_p}\\
Q_i &~= - \sum_{k=1}^N B_{i,k} = 0  & \text{ for the reactive power}
\end{align}
\end{enumerate}

\subsection{Computation of current flows from DC equations}
As we can see, the DC equations does not allow to capture flows in amps (A). Multiple methods allow to do that. We choose to do the following.

The AC equations \ref{eq:ac_i} gives us:
\[I_{i\to k} =~ \dfrac{\sqrt{P_{i\to k}^2+Q_{i\to k}^2}}{|V_i|}\]

We can obtained $P_{i\to k}$ from the DC equation \ref{eq:dc_p}, and $|V_i| = 1$ by assumptions. So the real problem is to compute $Q_{i\to k}$. Indeed, we can do better than simply assign $Q=0$ in these formulas.

\end{appendices}

\end{document}